# 从直接标记到压缩标记

## ——以《全宋文》墓志铭为例

**摘要** 本质上，直接标记和压缩标记都是利用自然语言表述中的固有规则来提取信息。直接标记有其优势，但在某些情况下却无法直接运用。比如在所提取数据本身并无规则而数据周边却是较为有规则的表述时，此时要先进行句子压缩和删除干扰信息然后在标记，就能够在不影响选取的前提下大幅提高所提取数据的精确度。

# From direct tagging to Tagging with sentences compression

## ——Taking Epitaphs in *Quansongwen* as an Example

**Abstract** In essence, the two tagging methods (direct tagging and tagging with sentences compression) are to tag the information we need by using regular expression which basing on the inherent language patterns of the natural language. Though it has many advantages in extracting regular data, Direct tagging is not applicable to some situations. if the data we need extract is not regular and its surrounding words are regular is relatively regular, then we can use information compression to cut the information we do not need before we tagging the data we need. In this way we can increase the precision of the data while not undermine the recall of the data.

## 1 问题的发现

目前中国历代人物传记资料库（Chinese Biographical Database，以下简称 CBDB）在建设数据库时主要采用的方法是标记（tagging）即先利用正则表达式提取文献中的信息，再通过人工审核补充标记的方法进行的。其主要流程为：第一步，首先整理出文献的电子文本，然后对文献进行一些简单的编辑处理以理清材料的出处、时间、作者和主题。第二步，明确所需要提取的具体信息。对于 CBDB 来说，人名、地名、官名、生卒年、亲属关系以及社会关系等是主要提取的对象。第三步，编辑正则表达式。而利用正则表达式提取信息，则需要熟悉相关信息的表达格式，做到用尽可能少的正则表达式提取尽可能多的信息。第四步，编写程序，将所需信息提取出并放入相应的表格中，同时在原文中标记出来，以便后期审核。第五步，人工校对，查漏补缺，完善数据格式。第六步，放入数据库中。当然，这些步骤之间并非毫无关联，而且它们之间的关联不仅仅存在于在前步骤对于在后步骤的影响，同时也存在于在后步骤对于在前步骤的影响，各步骤间既有“传导—决定”也有“反馈—改进”的关系。没有正则表达式而仅仅依靠人工提取信息会耗费大量时间、影响工作效率；而人工审核又会

反馈出之前正则表达式的不足，为修改正则表达式提供支持。

笔者在做《全宋文》墓志铭信息提取工作时就先尝试使用这个方法。不过在实践过程发现了一些问题，主要是提取人名中容易出现问题。在长辈姓名提取中这个问题并不明显，因为有许多关键词比如讳、曰、娶、氏和句尾标点等符号作为人名的提示，计算机能够识别出这些特征并进行提取。但是在子孙辈提取姓名时会出现计算机无法精确地识别人名的问题，比如下面这一句信息：

孫男二十人：長仲俶，右武衛大將軍、眉州刺史；次仲誘，右武衛大將軍、茂州刺史；次仲虒，右武衛大將軍、春州刺史；次仲礬，右監門衛大將軍；次仲沃、仲芮、仲雪、仲敌、仲靡，并右千牛衛將軍；次仲頎、仲霤、仲吟、仲疇、仲逢，并太子右監門率府率；次仲誥、仲譚、仲慥，并太子右內率府率；餘未命[1]。

无论用長、次还是分号等作为提示词或提示符号，都无法提取出所有的人名信息，将会有至少一半的人名信息被忽略，也就是说虽然发现了这些表达结构，但是却无法依据正则表达式准确的发现这些名字。

笔者仔细审视这段话所包含的信息发现，大量的信息是关于地名和官名的信息，如果剔除或者替代这些信息后，那么人名是否就容易被识别出来呢？笔者就尝试剔除了这些信息，于是这段话就变成了：

孫男二十人：長仲俶,、;次仲誘,、;次仲虒,、;次仲礬,;次仲沃、仲芮、仲雪、仲敌、仲靡，并；次仲頎、仲霤、仲吟、仲疇、仲逢，并；次仲誥、仲譚、仲慥，并；餘未命。

显然这就比之前更容易被提取出来，如果再对个别词汇作进一步处理，则就很容易提取出人名信息。

## 2 新方法的引入

上面这个例子，其实在告诉我们，人名信息虽然不是以有规律的方式呈现的，但人名周边的信息却是以某种有规则的方式出现，比如官名、地名、并，余等等。那么对于有规则的表达就可以利用计算机的工具进行处理，我们就可以把人名段落中的官名地名信息替换掉，实质上与正则表达式提取信息有异曲同工之妙，前者反向凸显人名信息，先删除有规则表达然后再提取人名，而后者根据人名的有规则表达直接提取人名信息。这种思路适合全部的墓志铭信息吗？

为了验证我们的方法是否对整个《全宋文》墓志铭有效，笔者进行了随机抽样（抽取全体中的百分之十），并进行了官名地名以及标点的替换。通过熟读关于孙男的墓志铭信息，我们得到“孫男.{0,2}人[,：][^女孫].{0,50}”这一正则表达式[5]。在 visual studio code 中 run，在全部墓志铭中可以发现 787 個相关結果。我们看一下用这一表达式能够发现的句子：

1，孫男二人：長應運，登丙戌進士第，儒林郎、兩浙轉運司物料官，即亨之也；次應龍，習舉子業[2]。

2，孫男五人：汝直、汝敦、汝平、汝功、汝能、皆業進士[3]。

3，孫男六人：曰夷仲，曰虞仲，曰於仲，曰南仲，曰武仲，曰延仲[4]。

4，孫男二十人：長仲俶，右武衛大將軍、眉州刺史；次仲誘，右武衛大將軍、茂州刺史；次仲虒，右武衛大將軍、春州刺史；次仲礬，右監門衛大將軍；次仲沃、仲芮、仲雪、仲敌、仲靡，并右千牛衛將軍；次仲頎、仲霤、仲吟、仲疇、仲逢，并太

子右監門率府率；次仲誥、仲謜、仲慥，并太子右內率府率；餘未命。

显然第三类是比较规则的句子，可以利用正则表达式直接提取，第一类和第三类经过后期程序编写也可以提取。第四类我们之前例子已经分析，是比较难以提取的，要经过比较复杂的编程才可以提取出人名。同一个正则表达式寻找到的结果在二次编程处理时却有着不同的难度，需要进行不同的处理。我们再看抽样的结果：

将新的正则表达式"孫男.{0,2}人([\w|\/]+([^孫女]{1,10})\/*)+"在 visual studio code 中 run, 选择全字匹配和正则表达式，可以提取 87 個結果，约等于从全体中提取结果（787）的 10%，这与上面的正则表达式提取结果基本一致。不过再排除官名、地名后，剩余的信息为人名的可能性很高，由此我们提高了数据的精确度。我们看一下上面四个句子在替换后的结果：

1，孫男二人/wm/長應運/wsep/登丙戌/no_noc/第/wsep//no_noc//wsep/兩浙[①]/no_noc//wsep/即亨之也/wsep/次應龍/wsep/習舉子業

2，孫男五人/wm/汝直/wsep/汝敦/wsep/汝平/wsep/汝功/wsep/汝能/wsep/皆/vno//no_noc/

3，孫男六人/wm/曰夷仲/wsep/曰虞仲/wsep/曰於仲/wsep/曰南仲/wsep/曰武仲/wsep/曰延仲

4，孫男二十人/wm/長仲僦/wsep//no_noc//wsep//ns//no_noc//wsep/次仲誘/wsep//no_noc//wsep//ns//no_noc//wsep/次仲虒/wsep//no_noc//wsep//ns//no_noc//wsep/次仲罊/wsep//no_noc//wsep/次仲沃/wsep/仲芮/wsep/仲雪/wsep/仲敔/wsep/仲靡/wsep/并/no_noc//wsep/次仲頎/wsep/仲霤/wsep/仲吟/wsep/仲巋/wsep/仲逄/wsep/并/no_noc//wsep/次仲誥/wsep/仲謜/wsep/仲慥/wsep/并/no_noc//wsep/餘未命

虽然也会出现不同于人名的其他词汇如"皆"、"曰"、"長"、"次"，但都比较规则，很容易编写程序进行排除，下文会进一步讨论这一问题。显然替代后提取会减少很多后期编程的复杂性，使得整个提取流程更加有效率。

## 3 新流程的探索

我们就尝试用这一方法去提取信息，改进 CBDB 的工作流程。根据我们对《全宋文》墓志铭的观察，几乎所有亲属关系的表述都在同一个句子中表达，这样就可以通过提取关键句子来确定信息提取的范围。因此新流程的第一步是将《全宋文》墓志铭中的含有亲属关系的句子提取出来。在这一步的编程中，要将墓志铭按照句子进行切分，并将所有的标点符号进行替换，为以后撰写正则表达式扫除不规则的表达。就需要做一个表，将《全宋文》墓志铭中出现的亲属关系都放入其中，并清楚地界定各个亲属关系的含义。我们基于 CBDB 原有的亲属关系表，进行了补充和编码，建立了新的关系表格（参见附表表 1song_kinship_title.xlsx）。除了"祖"、"考"、"子"、"男"[②]等直接表示亲属关系的词汇外，还需要将"娶"、"配"、"嫁"、"归"、"适"等表示婚姻关系的词汇收入其中，因为这些词间接指向女性亲属如祖妣或者男性亲属如女婿或者孙婿等。但在处理"配"、"归"、"适"这些词汇时要多加留意，因为它们是多义词，表示嫁娶只是它们众多含义中的一个，它们在墓志铭中可能以嫁娶之外的含义出现。因此我们在提取含有这些词汇的句子时，需要加上其他条件，即句子中必须同时包含

“氏”、“夫人”、“女”、“女孙”等词汇中至少一个时才能提取。这样我们就基本上将所有包含亲属关系的句子提取出来，消除了很多“噪音”，完成了工作的第一步。

第二步，压缩句子。将句子中的官名和地名分别替代为 no_noc 和 ns。这就需要制作官名表和地名表，CBDB 中有宋代的官名表和地名表，我们在这个基础上进行修改。必须指出，如果没有宋代的官名表和地名表，这项工作就会很受影响，整个新流程也将会变得不如之前流程方便、高效。这也是数据累计带来的设计上的多种可能性。在处理官名表和地名表时，需要注意尽可能不要出现一个字的官名或地名，比如“令”、“守”。但有些含义比较单一的词汇可以保留，比如表示通判的“倅”。需要特别指出的，有些地名或官名与亲属关系名称重合，这部分官名地名也需要考虑是否删除。比如长子这个地名，如果作为地名全部替换就会出现将大量的作为亲属关系名称的长子替换为 ns，会影响亲属信息的提取。此外表示亲属关系的庶子与表示官名的太子庶子，表示县名的卢氏与表示姓氏的卢氏虽然可能混淆，但这样的重合非常少，可以忽略。经过排查和重设，我们建立了官名和地名词典（分别见附表表 3song_no_dict.csv 和附表表 4song_ns_dict.csv）。此外，我们还建立了官职前常用委任词表等等（权，迁，授，赠等，见附表 5song_appt_type.xlsx）。在建立各种表之后，就要运行相应的替换，在编写程序时要注意他们的替换顺序，一般以字数长短为优先级，长字符串优先安排，同时地名表要优先于官名表。这样，我们就完成了句子的压缩。

第三步，编辑正则表达式。在有限的文本中，我们可以穷尽所有的表达式，但会非常耗时，且与后面审查的工作重复，得不偿失。因此，我们要用尽可能少的正则表达式提取尽可能多的所需信息。在 CBDB，一般而言用 20-30 个正则表达式提取百分之七十左右的信息，是一个理想的工作原则。那么在浩如烟海的资料中，如何确定哪些是这 20-30 个表达式，哪些不是呢？在流程改造前，笔者曾就亲属关系撰写了一百多个正则表达式，并标出提取数据的数量以及其中的一个示例（详见附表表 6 regular-expressions.xlsx，若无例子或计数，则可以直接在 visual studio code 中运行而得到。）。根据提取信息的数量进行排序[③]，并通过合并类似的表达式，就能基本确定哪些是我们需要的表达式。根据笔者的经验，一般情况下，如果一个表达式仅能提取 50 条以下的数据，基本上是没有必要撰写的。

第四步，编写程序。这里我们不讨论那些能够直接且容易提取到信息的正则表达式，因为不需要人文学者提出特殊的要求或建议。但在子孙、女、女婿等亲属提取方面需要人文学者的参与才能有效的提取到信息。我们还以前面的四个句子为例，看如何提取到有效信息。

1，孫男二人/wm/長應運/wsep/登丙戌/no_noc/第/wsep//no_noc//wsep/兩浙/no_noc//wsep/即亨之也/wsep/次應龍/wsep/習舉子業

2，孫男五人/wm/汝直/wsep/汝敦/wsep/汝平/wsep/汝功/wsep/汝能/wsep/皆/vno//no_noc/

3，孫男六人/wm/曰夷仲/wsep/曰虞仲/wsep/曰於仲/wsep/曰南仲/wsep/曰武仲/wsep/曰延仲

4，孫男二十人/wm/長仲俶/wsep//no_noc//wsep//ns//no_noc//wsep/次仲誘/wsep//no_noc//wsep//ns//no_noc//wsep/次仲虒/wsep//no_noc//wsep//ns//no_noc//wsep/次仲罃/wsep//no_noc//wsep/次仲沃/wsep/仲芮/wsep/仲雪/wsep/仲敔/wsep/仲靡/wsep/并/no_noc//wsep/次仲頎/wsep/仲霤/wsep/仲吟/wsep/仲疇/wsep/仲逄/wsep/并/no_noc//wsep/次仲誥/wsep/仲諲/wsep/仲慥/wsep/并/no_noc//wsep/餘未命

在地名和官名信息被替换后，我们发现还有一些干扰词汇，这些干扰词有很多事共通的，大概有一下几类[4]：

表示次序：長，次，幼，曰，季曰[5]，伯曰，仲曰，叔曰，長即，也，次即。

表示科举：貢，等，第，及第，中第，中舉，舉子，登，科。

表示就任官职的动词：今，今以，今爲，授，事，都，轄，新，知，舊，監，倉，庫，起，終，故，前，後，左，右。

表示行政区：州，軍，路，郡，縣，府。

表示地名：江淮，兩浙，寺。

表示官职：尉，某官，官，税務，支鹽。

表示社会身份：士族，士人。

提示职业的词汇：俱，業，業，習。

提示两人以上的词汇：皆，並，并，餘，俱，竝。

表示人生过程的词汇：未，未冠，未仕，未官，未命，先歿，先亡，先公，早夭，早亡，早世，夭，卒，幼，尚，尚幼，未名，前卒，先卒，蚤卒，俱有，早，早卒，喪，早喪，

表示仕宦：未仕，未銓，左銓，司户，户部，户。

其他固定搭配：一，一人，二，二人，一尚，二尚，三尚，三，三人，三曰，四人等。

由于计算机无法识别哪些是表示名字的词汇，所以需要把这些词汇在编程中进行批量处理。这一点和前面的替换在原理上是一致的。把这些词汇删除之后，我们再看看这四个句子的情况：

1，孫男二人/wm/應運/wsep/丙戌/no_noc//wsep//no_noc//wsep//no_noc//wsep/即亨之/wsep/應龍/wsep/

2，孫男五人/wm/汝直/wsep/汝敦/wsep/汝平/wsep/汝功/wsep/汝能/wsep//vno//no_noc/

3，孫男六人/wm/夷仲/wsep/虞仲/wsep/於仲/wsep/南仲/wsep/武仲/wsep/延仲

4 孫男二十人/wm/仲俶/wsep//no_noc//wsep//ns//no_noc//wsep/仲誘/wsep//no_noc//wsep//ns//no_noc//wsep/仲颸/wsep//no_noc//wsep//ns//no_noc//wsep/仲謍/wsep//no_noc//wsep/仲沃/wsep/仲芮/wsep/仲雪/wsep/仲敔/wsep/仲靡/wsep//no_noc//wsep/仲頎/wsep/仲霤/wsep/仲吟/wsep/仲疇/wsep/仲逢/wsep//no_noc//wsep/仲誥/wsep/仲譚/wsep/仲慥/wsep//no_noc//wsep/

在删除干扰词汇后，除了第一个句子中的“丙戌”被误认为名字外，“即亨之”由于超过两个字而不会被认为名字，因为这里的名字略去了姓，而古代的人名极少出现一个姓氏后面加三个字的情形。所有其他剩下的词汇都是人名了。而正则表达式的功能就不在是精确提取人名，而是在程序中被当作亲属信息的结构提示了。那么这些表示子孙的正则表达式相比于未替换或未删除干扰信息就简化了很多，最主要的表达式如下：

[^女]*(曾孫|孫)([^男女]{0,3}?)人(\/.+)

[^女]*(孫子|子)([^男女]{0,3}?)人(\/.+)

(子|孙|孫|曾孫)男(.{0,3}?)人(\/.+)

[^子孫孙]男(.{0,3}?)人.{0,50}

(子|孫|曾孫|曾孫男|男|孫男|子男|孙男|孙)[一二三四五六七八九十]/

(子|孫|曾孫|曾孫男|男|孫男|子男|孙男|孙|女|婿)曰[^/].{0,50}

生[一二三四五六七八九十][男|子][^女孫].{0,150}

[^第][一二三四五六七八九十][男|子][^女孫].{0,150}?/wm/

以上几个正则表达式帮助我们找到 4000 条左右的子孙信息，如果按照每个信息有三个人名，那么将会得到 12000 条亲属信息。关于女儿女婿的信息也可以通过类似的方法提取，这里就不赘述。

第五步，人工审核。与此前审核需要通看全文，查漏补缺不同，句子压缩后，亲属信息的百分之 99 以上的信息应该都在压缩的句子中，在压缩句子中就已经将众多的信息删除，方便人工审核时快速找到信息。在输出结果形式上，我们采用 Excel 表格，便于批量删除，大大提高了人工审核阶段的效率。由于能在 Excel 表格中直观地看到大量的数据，也便于发现存在问题的规律，有利于反馈给编程人员关于正则表达式的修正方案或者干扰词汇的删除工作。

第六步，数据入库，这部分工作主要属于编程人员。

## 4 结语

无论哪种标记方法本质上都是利用自然语言表述中的固有规则来提取我们所需要的信息。经过句子压缩，我们节省了很多时间和成本，更加高效地利用正则表达式标记我们所需要的信息。正则表达式也不在仅限于标记一些表述简单的信息，扩大了其应用的范围。同时在某种程度上，改变了其本来的意义，即不是作为准确信息提取的工具而是成了提示信息出现的工具，与编程相配合，高效地解决了复杂表述中的信息标记问题。

## 注释

① 由于宋代地名表中没有两浙，在这里就没有被替换。又因这样的例子很少，因此没有必要添入地名表进行替换，下文在词典部分会进一步讨论这一问题。

②有必要指出，考、子、男等除表达亲属关系之外还有其他意思，尤其是子，在古代还有个常用含义是先生老师的尊称，由此引申，一些伟大的思想家在后世就以姓氏加上“子”的形式被尊称，如孔子、孟子、老子和庄子。因此我们还要做一个仔细的审查与排除，将这些专有名词删除。还有一类常见的亲属关系表述是两个不同亲属连称，如父母、父子、男女、兄弟，它们所在的句子往往并没有我们需要的信息，因此也要排查与删除，以减少后期审查工作的强度。最后一种常见的是亲属关系言说某些话，比如“母曰：”而非“母曰+名字”，这类表述也没有我们想要提取的信息，也需要排除。为此，我们建立了排除与删除表（参见附表表 2 字典排查與删除.xlsx）以供后期编程参考。

③需要注意的是，提取信息的量有时候虽然很多，但有效信息并不多。这需要平衡 precision 与 recall 之间的矛盾，根据笔者的经验，precision 达到百分之八十左右，可以作为一个标准，否则需要调整表达式结构，或者与其他表达式合并，或者取消该表达式。否则，由此带来的后期审查会非常麻烦，还不如让审查人员直接读原文标记信息。

④这些词汇首先根据正则表达式提取的文本先总结出大部分，然后运行编程得出结果

再进行进一步的修正。

⑤伯仲叔季四个字的每个字都不能单独删除，因为古代中国的名字中不少会出现伯仲叔季，比如第三和第四个句子就出现了“仲”，并且出现了 23 次之多。但“伯曰”、“仲曰”表示长幼次序的可以删除而不影响结果。这个例子再次提醒我们在筛选删除字词时必须慎重。

## 参考文献

## 附表（限于篇幅，各表仅显示部分）：

表 1

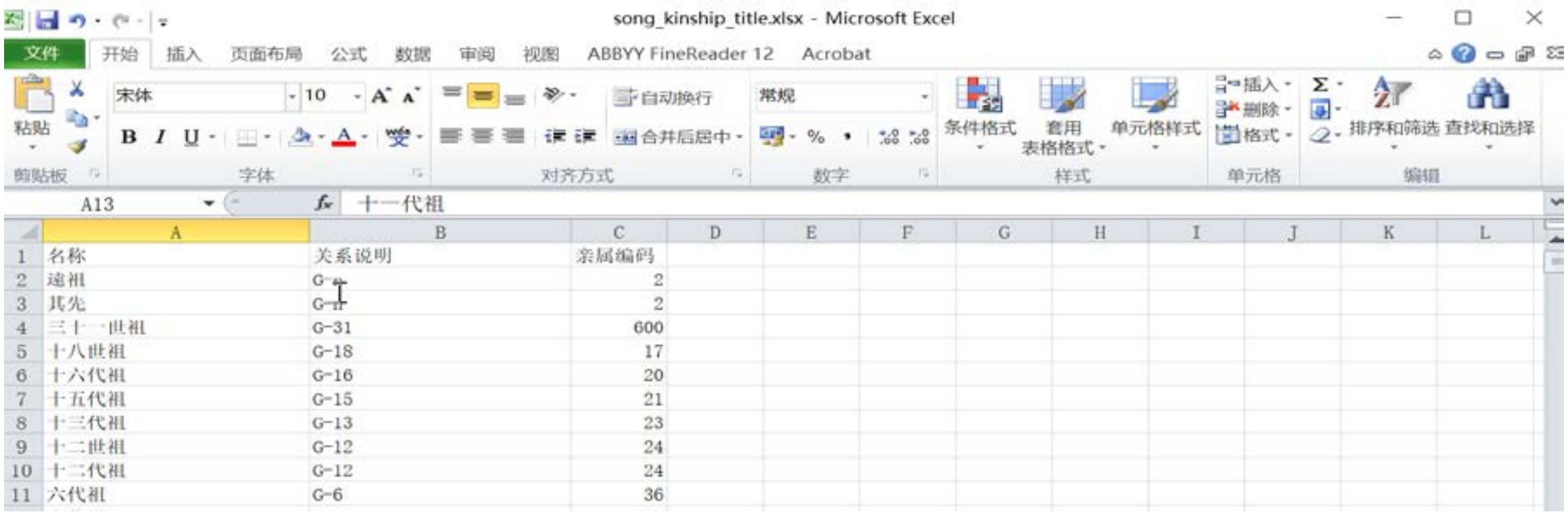

song_kinship_title.xlsx - Microsoft Excel

A13 十一代祖

| | A | B | C |
|---|---|---|---|
| 1 | 名称 | 关系说明 | 亲属编码 |
| 2 | 遠祖 | G-[illegible] | 2 |
| 3 | 其先 | G-[illegible] | 2 |
| 4 | 三十一世祖 | G-31 | 600 |
| 5 | 十八世祖 | G-18 | 17 |
| 6 | 十六代祖 | G-16 | 20 |
| 7 | 十五代祖 | G-15 | 21 |
| 8 | 十三代祖 | G-13 | 23 |
| 9 | 十二世祖 | G-12 | 24 |
| 10 | 十二代祖 | G-12 | 24 |
| 11 | 六代祖 | G-6 | 36 |

表 2

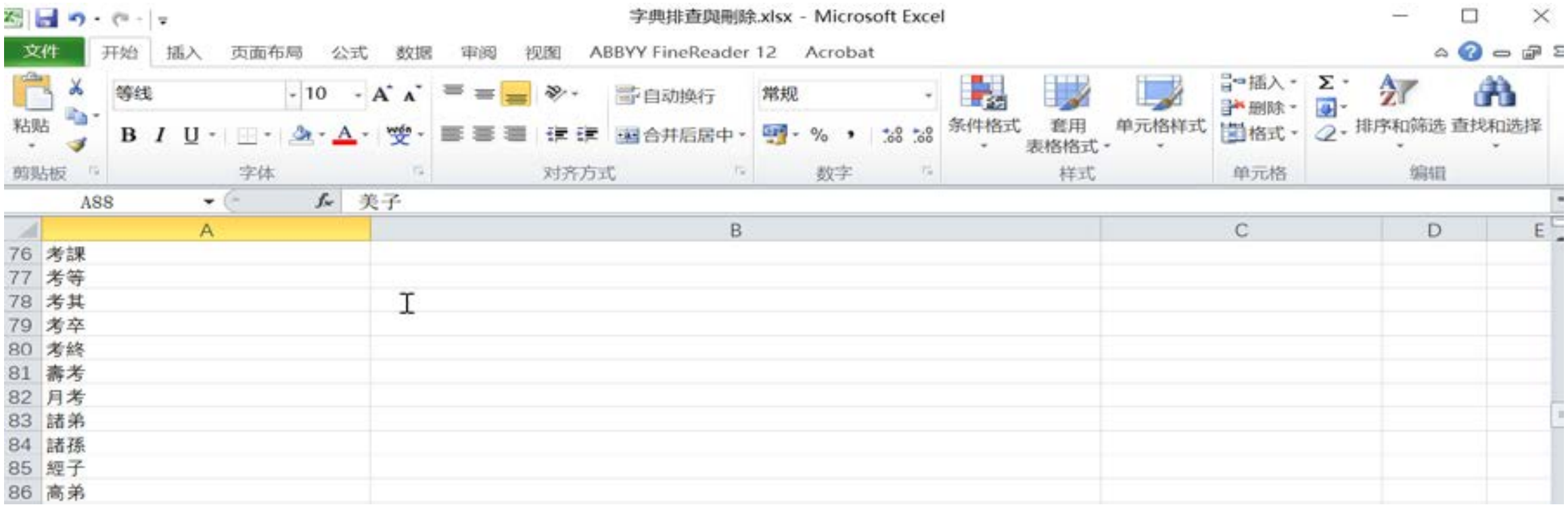

字典排查與刪除.xlsx - Microsoft Excel

A88 美子

| | A |
|---|---|
| 76 | 考課 |
| 77 | 考等 |
| 78 | 考其 |
| 79 | 考卒 |
| 80 | 考終 |
| 81 | 壽考 |
| 82 | 月考 |
| 83 | 諸弟 |
| 84 | 諸孫 |
| 85 | 經子 |
| 86 | 高弟 |

表 3

| | | |
|---|---|---|
| 2 | 同管勾成都府利州陝西等路茶事兼提舉陝西等路買馬監牧公事 | no |
| 3 | 都大提舉成都府利州等路茶事兼提舉四川等路買馬監牧公事 | no |
| 4 | 都大提舉成都府利州陝西等路兼提舉陝西等路買馬公事 | no |
| 5 | 總領四川財賦軍馬錢糧所幹辦行在分差戶部魚關糧料院 | no |
| 6 | 總領四川財賦軍馬錢糧所幹辦行在分差戶部利州糧料院 | no |
| 7 | 都大提舉成都府利州陝西等路茶事司兼提舉陝西買馬 | no |
| 8 | 總領湖廣江西京西路財賦湖北京西軍馬錢糧所 | no |
| 9 | 秦鳳路馬步軍都部署兼經略安撫沿邊招討使 | no |
| 10 | 涇原路馬步軍都部署兼經略安撫沿邊招討使 | no |
| 11 | 鄜延路馬步軍都部署兼經略安撫沿邊招討使 | n[illegible] |
| 12 | 環慶路馬步軍都部署兼經略安撫沿邊招討使 | no |
| 13 | 淮南江浙荊湖制置發運使兼經制兩浙江東路 | no |
| 14 | 淮南江浙荊湖制置茶鹽礬稅兼都大發運都监 | no |
| 15 | 都大提舉成都府利州秦鳳熙河等路茶場司 | no |
| 16 | 陝西馬步軍都部署兼經略安撫沿邊招討使 | no |
| 17 | 淮南江浙荊湖制置茶鹽礬稅兼都大發運使 | no |

表 4

| | | |
|---|---|---|
| 557 | 吳興郡 | ns |
| 558 | 新興郡 | ns |
| 559 | 嘉興郡 | ns |
| 560 | 伏羌城 | ns |
| 561 | 德陽郡 | ns |
| 562 | 博平郡 | ns |
| 563 | 濟陽郡 | ns |
| 564 | 樂安郡 | ns |
| 565 | 江夏郡 | ns |
| 566 | 鉅鹿郡 | ns |
| 567 | 恩平郡 | ns |
| 568 | 甘谷城 | ns |
| 569 | 開封府 | ns |
| 570 | 應天府 | ns |
| 571 | 河北道 | ns |
| 572 | 秦鳳路 | ns |
| 573 | 廣濟軍 | ns |
| 574 | 永靜軍 | ns |

表 5

| | A | B | C | D | E |
|---|---|---|---|---|---|
| 3 | 1 | 權 | Provisional Appointment | | |
| 4 | 2 | 行 | Lower Acting Appointment | | |
| 5 | 3 | 守 | Higher Acting Appointment | | |
| 6 | 4 | 試 | Exceptional Acting Appointment | | |
| 7 | 5 | 權發遣 | Temporary Appontment | | |
| 8 | 6 | 借 | Temporary Higher Office Appointment | | |
| 9 | 7 | 假 | Temporary | | |
| 10 | 8 | 添差 | Extraordinary Appointment | | |

表 6

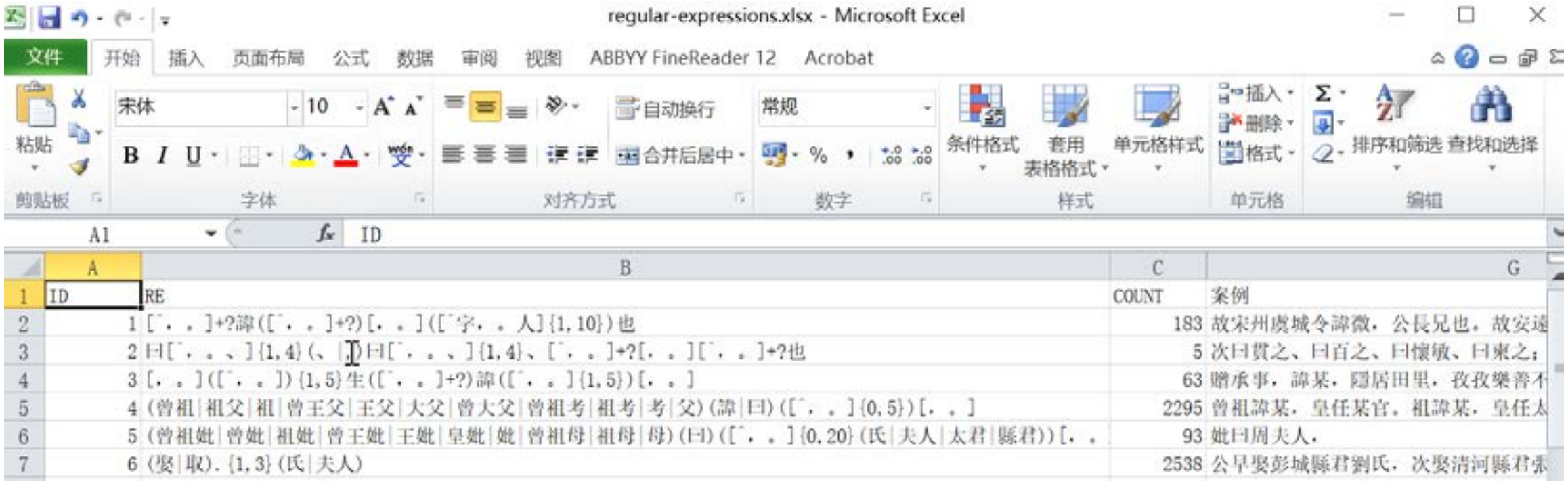

<table>
<tr><th>ID</th><th>RE</th><th>COUNT</th><th>案例</th></tr>
<tr><td>1</td><td>[^，。]+?諱([^，。]+?)[，。]([^字，。人]{1,10})也</td><td>183</td><td>故宋州虞城令諱微，公長兄也，故安遠</td></tr>
<tr><td>2</td><td>曰[^，。、]{1,4}(、|，)曰[^，。、]{1,4}、[^，。]+?[，。][^，。]+?也</td><td>5</td><td>次曰貫之、曰百之、曰懷敏、曰東之；</td></tr>
<tr><td>3</td><td>[，。]([^，。]){1,5}生([^，。]+?)諱([^，。]{1,5})[，。]</td><td>63</td><td>贈承事，諱某，隱居田里，孜孜樂善不</td></tr>
<tr><td>4</td><td>(曾祖|祖父|祖|曾王父|王父|大父|曾大父|曾祖考|祖考|考|父)(諱|曰)([^，。]{0,5})[，。]</td><td>2295</td><td>曾祖諱某，皇任某官。祖諱某，皇任太</td></tr>
<tr><td>5</td><td>(曾祖妣|曾妣|祖妣|曾王妣|王妣|皇妣|妣|曾祖母|祖母|母)(曰)([^，。]{0,20}(氏|夫人|太君|縣君))[，。</td><td>93</td><td>妣曰周夫人。</td></tr>
<tr><td>6</td><td>(娶|取).{1,3}(氏|夫人)</td><td>2538</td><td>公早娶彭城縣君劉氏，次娶清河縣君張</td></tr>
</table>